\documentclass[10pt,twocolumn,letterpaper]{article}

\usepackage{wacv}
\usepackage{times, soul}
\usepackage{epsfig}
\usepackage{graphicx}
\usepackage{amsmath}
\usepackage{amssymb}
\usepackage{amsmath} 
\usepackage{mathtools}
\usepackage{graphicx}
\usepackage{multirow}
\usepackage{subfig}
\usepackage{comment}
\usepackage{authblk}
\usepackage[accsupp]{axessibility}
\newcommand\numberthis{\addtocounter{equation}{1}\tag{\theequation}}
\usepackage{mathtools}

\def\wacvPaperID{1267} 

\wacvalgorithmstrack   

\wacvfinalcopy 

\ifwacvfinal
\usepackage[breaklinks=true,bookmarks=false]{hyperref}
\else
\usepackage[pagebackref=true,breaklinks=true,colorlinks,bookmarks=false]{hyperref}
\fi

\begin{document}

\title{CFL-Net: Image Forgery Localization Using Contrastive Learning}

\author[$\dag$]{Fahim Faisal Niloy}
\author[$\ddag$]{Kishor Kumar Bhaumik}
\author[$\ddag$]{Simon S. Woo}

\affil[$\dag$]{Center for Computational \& Data Sciences, Independent University, Bangladesh}
\affil[$\ddag$]{Computer Science and Engineering Department, Sungkyunkwan University, Suwon, South Korea}

\affil[ ]{{\tt\small niloy9542@gmail.com, \{kishor25, swoo\}@g.skku.edu}}

\renewcommand\Authands{ and } 


\maketitle
\thispagestyle{empty}

\begin{abstract}
   Conventional forgery localizing methods usually rely on different forgery footprints such as JPEG artifacts, edge inconsistency, camera noise, etc., with cross-entropy loss to locate manipulated regions. However, these methods have the disadvantage of over-fitting and focusing on only a few specific forgery footprints. On the other hand, real-life manipulated images are generated via a wide variety of forgery operations and thus, leave behind a wide variety of forgery footprints. Therefore, we need a more general approach for image forgery localization that can work well on a variety of forgery conditions. A key assumption in underlying forged region localization is that there remains a difference of feature distribution between untampered and manipulated regions in each forged image sample, irrespective of the forgery type. In this paper, we aim to leverage this difference of feature distribution to aid in image forgery localization. Specifically, we use contrastive loss to learn mapping into a feature space where the features between untampered and manipulated regions are well-separated for each image. Also, our method has the advantage of localizing manipulated region without requiring any prior knowledge or assumption about the forgery type. We demonstrate that our work outperforms several existing methods on three benchmark image manipulation datasets. Code is available at \href{https://github.com/niloy193/CFLNet}{https://github.com/niloy193/CFLNet}
\end{abstract}

\section{Introduction}
Image forgery has been a serious emerging socio-technical issue, as more advanced AI techniques have been leveraged to create fake images. Image is a significant medium for information transfer. In order to produce fake stories, academic trickery, and illegal conduct, manipulated photographs created utilizing image editing technology are constantly being mistaken for real ones. When a digital image is manipulated, we frequently assume that image forensic investigations will be able to spot the tampered areas. However, collecting differentiating features of tampered areas with various forging types (including splicing, copy-move, removal, etc.) is still challenging and typically calls for utilizing the special qualities of numerous tampering artifacts.


\begin{figure}
\setlength\tabcolsep{1pt}
\centering
\begin{tabular}{ccc}
 \includegraphics[width=1in]{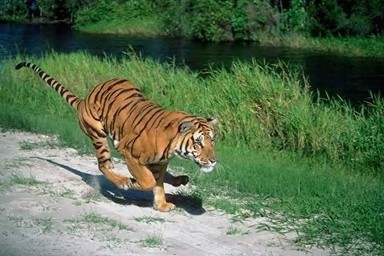} &
 \includegraphics[width=1in]{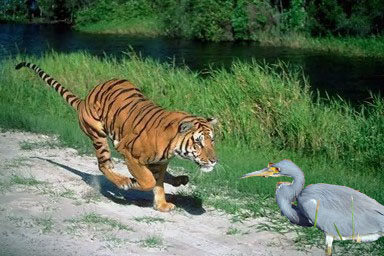} &
 \includegraphics[width=1in]{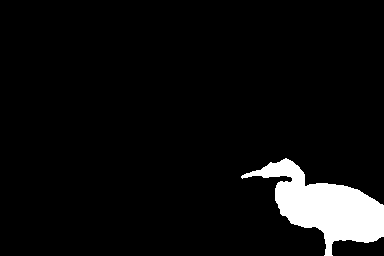} \\
\includegraphics[width=1in]{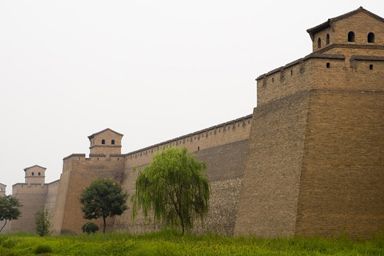} &
 \includegraphics[width=1in]{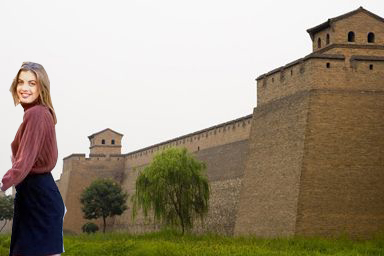} &
 \includegraphics[width=1in]{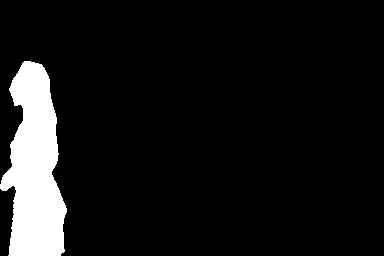} \\
 \includegraphics[width=1in]{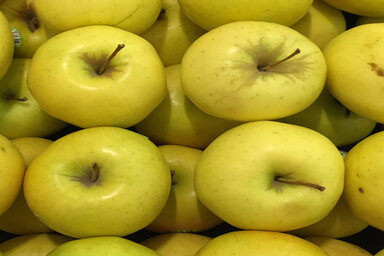} &
 \includegraphics[width=1in]{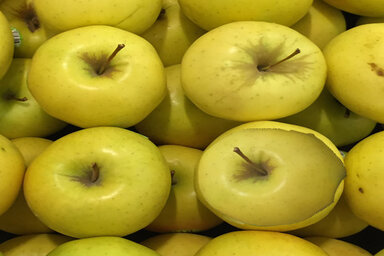} &
 \includegraphics[width=1in]{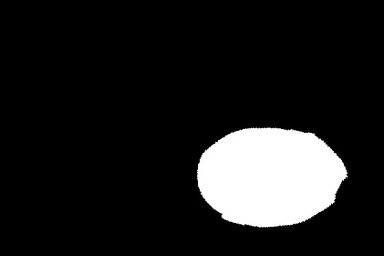} \\
 \includegraphics[width=1in]{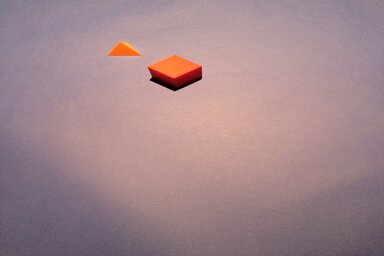} &
 \includegraphics[width=1in]{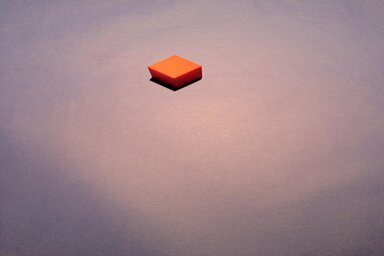} &
 \includegraphics[width=1in]{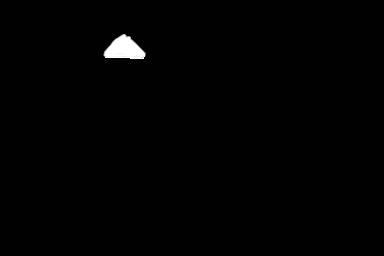} \\
 \textbf{\footnotesize{Original Image}} &
 \textbf{\footnotesize{Manipulated Image}} &
 \textbf{\footnotesize{Ground Truth Mask}} \\
\end{tabular}
\caption{Examples of image manipulation. First two rows show examples of image splicing and the next two rows show examples of copy-move forgery and removal respectively.}
\label{fig:intro}
\end{figure}

Generally, image forgery can be broadly categorized into: splicing \cite{cozzolino2015splicebuster, kniaz2019point}, copy-move \cite{cozzolino2015efficient,wu2018image, wu2018busternet}, removal \cite{zhu2018deep}, enhancement \cite{bayar2018constrained, choi2017detecting}, etc.  First, in image splicing, content is copied and pasted from other source images, as opposed to copy-move forgery, where the content is obtained from the same image. On the other hand, removal or inpainting techniques remove a selected region from the image and fills the space with new pixel values estimated from background \cite{wu2019mantra}. Image enhancement exploits a wide collection of local manipulations, such as sharpening, brightness adjustment, etc. Each of the broader categories can be further divided into more fine-grained forgery types. For example, Gaussian blurring or JPEG compression may be applied to the tampered region before committing splicing or copy-move forgery. Recently, more general-purpose image forgery localization methods have been proposed, which can detect or localize more than one forgery type, such as RGB-N Net \cite{zhou2018learning}, Manipulation Tracing Network (ManTraNet) \cite{wu2019mantra}, Spatial Pyramid Attention Network (SPAN) \cite{hu2020span}, etc.

These general image forgery detection or localization methods usually rely on different forgery clues or footprints left by the forgery operation, such as JPEG artifacts \cite{li2017image, amerini2017localization}, edge inconsistency \cite{salloum2018image, zhang2018boundary}, noise pattern \cite{cozzolino2019noiseprint, yang2016image}, camera model \cite{rafi2020l2}, EXIF inconsistency \cite{huh2018fighting}, etc., to detect or localize forgery. Table 1 of \cite{wu2019mantra} summarizes existing major forgery localization methods and the forgery clues the methods focus on. For example, \cite{bappy2017exploiting} employs LSTM based patch comparison to focus on edge inconsistency between the tampered patches and authentic patches. CAT-Net \cite{kwon2021cat} leverages DCT coefficients to focus on resampling clues. 

However, training models to focus on specific forgery clues has a major disadvantage. Because then, the model can only detect forgery if that particular forgery footprint is prominent in the forged image. This is unacceptable because, in real-life, different manipulation techniques can leave behind wide variety of forgery clues. Thus, focusing on specific forgery clues is not optimal. For example, if a method focuses on edge inconsistency to detect forgery, the method will not perform well on a forged image where the boundary between untampered and manipulated region is smooth. Again, if a method focuses on resampling features, it will struggle to detect forgery if an image has the same JPEG compression applied several times to both the untampered and manipulated regions. 

Another major disadvantage of existing methods is that these methods use cross-entropy loss without additional constraints for training. Recently, \cite{zhao2020makes} stated that traditional cross-entropy based methods assume that all instances within each category should be close in feature distribution. This ignores the unique information of each sample. Thus, cross-entropy loss encourages the model to extract similar features for same category. This might be helpful for classification or segmentation of datasets such as Imagenet or Cityscapes, where objects of the same category should have similar features. However, in the case of image forgery localization, extracting similar features for all the tampered regions in the dataset is not optimal as different manipulation operations leave behind different forgery footprints in the tampered regions. Hence, without additional constraints, a common cross-entropy loss-based framework is prone to over-fitting on specific forgery patterns \cite{luo2021generalizing}. This is not conducive to generalization.


Taking all these limitations into consideration, we propose a novel forgery localization method named \textit{Contrastive Forgery Localization Network} or \textit{CFL-Net}, based on recently proposed contrastive loss \cite{khosla2020supervised}. Our method relies on the general assumption in underlying forged region localization that there remains a difference of feature statistics, i.e., color, intensity, noise, etc., between untampered region and manipulated region \cite{hu2020span}, irrespective of the forgery type. In this paper, we focus on leveraging this difference in the feature space to aid in image forgery localization via contrastive loss. Specifically, our model learns mapping into a feature space where the features between untampered and manipulated regions are well-separated and dispersed for each image. Thus, our method does not focus on specific forgery clues. Also, we calculate the contrastive loss for each sample. Hence, our method treats the forgery clues of each sample differently, which helps in generalization. Our main contributions are summarized as follows:

\begin{itemize}
    \item We propose a novel image forgery localization method called \textit{CFL-Net}. Our method leverages the difference of feature distribution between untampered and manipulated regions of each image sample and does not focus on specific forgery footprints. Hence, our method is more well-suited to detect real-life forgery.
    
    \item We address the problem of using cross-entropy loss without any constraints for general purpose image forgery localization. We incorporate contrastive loss and especially tailor it towards solving this problem.
    
    \item We perform extensive experiments on benchmark manipulation datasets to show that our method outperforms several existing image forgery localization methods.
\end{itemize}

\section{Related Works}

\begin{figure*}[!t]

 \centering
  \includegraphics[width=1\textwidth]{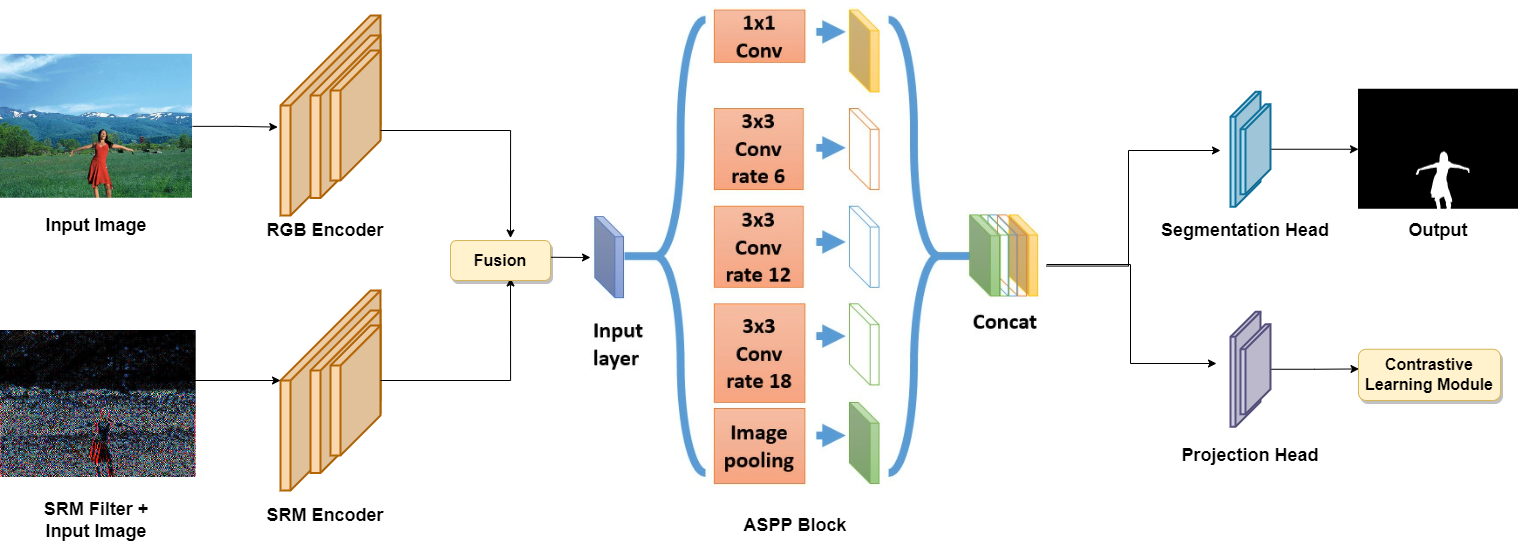}
 \caption{Overall architecture of the proposed CFL-Net. We use a two stream encoder, one for the RGB input image and the other for the SRM filtered image. The features produced by the encoders are fused and passed into the ASPP module. The output features from ASPP block then go through both the Segmentation Head and Projection Head, where the first produces the final prediction mask and the latter produces features that go into the contrastive learning module.}
 \label{fig:cfl}
\end{figure*}

\subsection{Image Forgery Localization}
Image forgery methods are concerned with forgery classification or localization. Classification is basically predicting whether an image is forged or non-forged, whereas, forgery localization is concerned with locating the forged region as well. The latter is a segmentation task.

In pre deep learning era, methods used hand-crafted features such as local noise analysis \cite{fridrich2012rich, cozzolino2014image}, CFA artifacts \cite{ferrara2012image}, JPEG compression \cite{bianchi2011improved} etc. Recent works usually use deep learning based methods in conjunction with these forgery traces to localize forged regions. Bappy et al. \cite{bappy2017exploiting} exploit the edge inconsistency trace using LSTM to localize forgery. The work is later improved in \cite{bappy2019hybrid}, where the authors further exploit resampling traces using Laplacian filters. They also use a separate encoder-decoder structure to refine the predicted mask. RGB-N \cite{zhou2018learning} proposes a two stream faster R-CNN network, one for the RGB image and other for the noise information traces generated by the Steganalysis Rich Model (SRM) filters \cite{fridrich2012rich}. SRM filters are high pass filters that enhance the high-frequency information, which becomes helpful in forgery localization. However, due to the R-CNN architecture, RGB-N is limited to localizing to a rectangular box whereas real objects are not necessarily rectangular. Mantra-Net \cite{wu2019mantra} jointly detects and localizes forged images. ManTra-Net is composed of a VGG based feature extractor and an LSTM based detection module. The feature extractor is trained to detect various types of image manipulation traces. SPAN \cite{hu2020span} proposes Spatial Pyramid Attention Network models the relationship between image patches at multiple scales by constructing a pyramid of local self-attention blocks. CAT-Net \cite{kwon2021cat} uses two stream network similar to RGB-N, one for the RGB pixel stream and the other for DCT co-efficients. DCT helps to extract resampling trace features.

\subsection{Contrastive Learning}
Recently, contrastive learning \cite{he2020momentum, chen2020simple} has achieved great progress
in unsupervised learning problem. SimCLR \cite{chen2020simple} proposes a simple framework to perform contrastive learning, where positive pairs are generated with two random augmented views of the same image and negative ones are obtained with different images, forming an image-level discrimination task. Furthermore, MoCo \cite{he2020momentum} maintains a queue of negative samples and turns one branch of Siamese network into a momentum encoder to improve consistency of the queue. Recently \cite{khosla2020supervised} has extended unsupervised contrastive learning to fully-supervised setting that can effectively leverage label information. This setting has been used in semantic segmentation to improve the state-of-the-art performance. \cite{wang2021exploring, hu2021region} contrast the pixel embedding between different semantic categories in a supervised manner to aid in segmentation. 

Sun et al. \cite{sun2021dual} have also used supervised contrastive loss to supplement cross-entropy loss for forgery detection task. However, their work is targeted toward forged face image classification. In contrast, our method is aimed toward general-purpose image forgery localization, which is a segmentation task. Also, the formation of our contrastive loss is different. Fung et al. \cite{fung2021deepfakeucl} use unsupervised contrastive learning for deepfake face image forgery detection. This method is also aimed toward only forgery classification.  

\section{CFL-Net}

\begin{figure*}[t!]

 \centering
  \includegraphics[width=1\textwidth]{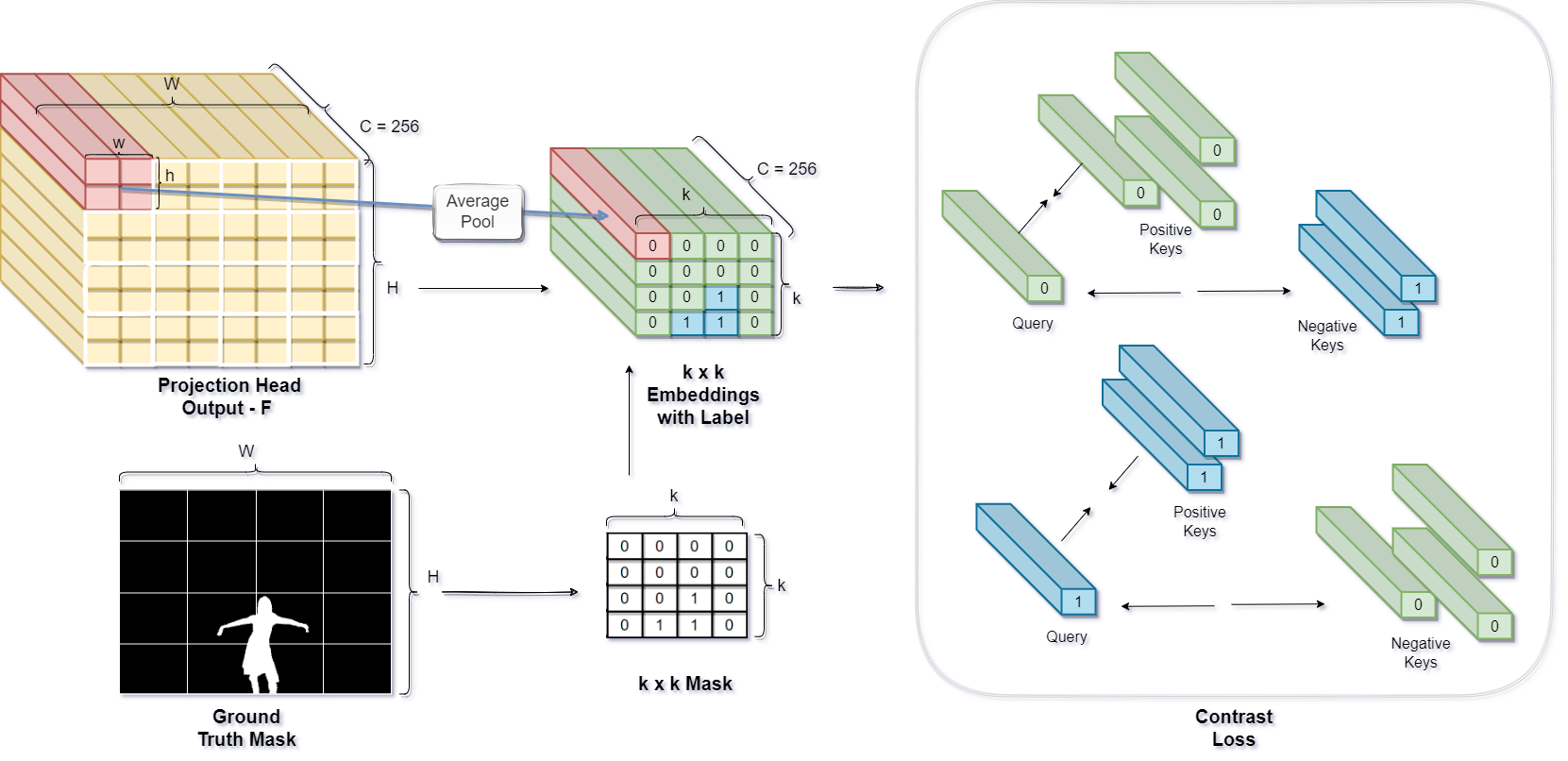}
 \caption{Contrastive Learning Module: For ease of visualization, the projection head in the figure is shown to output a feature map $F$ of shape $256 \times 8 \times 8$. The feature map is then divided into $4 \times 4$ patches. Then, all the $4$ spatial vectors in each patch are averaged to get the embeddings of size $4 \times 4$ (denoted as '$k \times k$ Embeddings with Label' in figure). The ground truth mask is also divided into $4 \times 4$ patches and maximum occurring pixel label in each patch is counted to get the output  $4 \times 4$ mask (denoted as '$k \times k$ Mask' in figure). Eqn. \eqref{con_loss_final} is then used to calculate the contrastive loss for each pixel embedding of the '$k \times k$ Embeddings with Label'.}
 \label{fig:con_loss}
\end{figure*}

In this section, we first describe the overall framework of our model. We then detail on the contrastive learning part.
\subsection{Overall Framework}
We present here the overall framework of our method. The overall diagram is shown in Figure \ref{fig:cfl}. We opt for a two stream network similar to \cite{zhou2018learning, kwon2021cat, sun2021dual}. One stream takes the input RGB image $I \in R^{3 \times H \times W}$ as input. We use SRM filters \cite{fridrich2012rich} to the RGB image and use that as an input for the other stream. SRM filters are high pass filters that enhance the high-frequency information of input image, thus highlighting the edge information more, which is helpful for localizing forgeries. We use ResNet \cite{he2016deep} as the backbone for both the streams. We then fuse features from both streams by concatenating features channel-wise. ASPP module \cite{chen2017rethinking} is used on the fused feature map so that multi-scale information can be extracted. It is reported in \cite{zhou2018learning} that global context helps to collect more clues, such as contrast difference, etc., for manipulation detection. ASPP module helps in this regard by extracting information in different scales, such that global context as well as more fine-grained pixel level context information becomes available.

We then use a segmentation head/decoder head and a projection head that takes the upsampled multi-scale feature extracted by the ASPP module as input. We opt for a DeepLab style segmentation head which outputs the final segmentation map of size $H \times W$. The projection map is composed of Conv-BatchNorm-Conv layer that projects the feature map to $F \in R^{256 \times H \times W}$, 256 being the embedding dimension. The embedded feature map $F$ is passed on to the contrastive learning module. The projection head is not used during evaluation. 

\subsection{Contrastive Learning Module}

Our goal is to contrast between the untampered and manipulated pixel embeddings of each sample so that the feature distributions between both regions get well-separated. As our embedded feature map is of size $H \times W$ spatially and we have the corresponding ground-truth mask $M$ of similar size, we know the label of each pixel embedding. Thus, we can use supervised contrastive learning. 
For each query pixel embedding $z_{i}$ the contrastive loss for that embedding becomes:

\begin{align*}
L_{i} {=} \frac{1}{|A_{i}|}{\sum_{k^{+} \in A_{i}}} {-\log{\frac{ exp(z_{i} {\cdot} k^{+} /\tau)}{exp(z_{i} {\cdot} k^{+} /\tau) + \sum_{k_{-}} exp(z_{i} {\cdot} k^{-} /\tau)}}} \numberthis \label{con_loss}
\end{align*}

Here, $k^{+}$ or positive key is a pixel embedding that has the same label as query $z_{i}$. $A_{i}$ denotes the set of all $k^{+}$ in the projection head output feature map $F$. Similarly, $k^{-}$ or negative key, are pixel embeddings in $F$ that have a different label than $z_{i}$.

However, calculating $L_{i}$ in such a manner has some major limitations. First, calculating the contrastive loss based on single pixel embedding does not take into account the context information that the neighboring embeddings have. Also, to calculate the loss, a dot-product matrix of size $HW \times HW$ needs to be stored, which is memory-consuming. 

One possible solution is to randomly sample a few pixel embeddings from $F$ corresponding to the two different classes similar to \cite{wang2021exploring}. Then, use those embeddings to calculate \eqref{con_loss}. This way, the memory requirement is greatly reduced. However, this solution does not take into account the context information from neighboring pixels. Also, similar to \cite{hu2021region}, another solution could be to average all the pixel embeddings of the two regions and then use the mean embeddings to calculate the loss. Although this may be helpful for computer vision tasks, such as segmentation of semantic objects etc., it is inappropriate for image manipulation detection tasks. Because, recent studies have shown that pooling is undesirable for tasks that require subtle signals since pooling reinforces content and suppresses noise-like signals \cite{boroumand2018deep}. These fine-grained traces are helpful for detecting forgery. Hence, to find a balance between context and fine-grained traces, we opt for dividing $F$ into local regions. 

We first partition $F$ spatially into $k \times k$ patches, thus getting $f_{i} \in  R^{256 \times h \times w}$, where $i \in \{1,2,3 ... k^{2}\}$ and $h = \frac{H}{k}$ and $w = \frac{W}{k}$. We then take the average of the pixel embeddings in each local region. Thus making each $f_{i}$ to a shape of $R^{256}$. In a similar manner, we divide the ground truth mask $M$ into $k \times k$ patches. $M$ has value of 0 in the untampered region and value of 1 in the forged region. We get $m_{i} \in  R^{h \times w}$, where $i \in \{1,2,3 ... k^{2}\}$ and $h = \frac{H}{k}$ and $w = \frac{W}{k}$. To get the value of the label of each $m_{i}$, we count the number of 0s and 1s in the $h \times w$ patch. We then assign the value of $m_{i}$ as the maximum count of value occurring in the patch. 

Now, we have pixel embeddings $f_{i}$ and corresponding label of each embedding $m_{i}$. We can now use the supervised contrastive loss as:

\begin{align*}
L_{i} {=} \frac{1}{|A_{i}|}{\sum_{k^{+} \in A_{i}}} {-\log{\frac{ exp(f_{i} {\cdot} k^{+} /\tau)}{exp(f_{i} {\cdot} k^{+} /\tau) + \sum_{k_{-}} exp(f_{i} {\cdot} k^{-} /\tau)}}} \numberthis \label{con_loss_final}
\end{align*}

Here also, $A_{i}$ denotes the set of all other pixel embeddings $k^{+}$ that have the same label as $f_{i}$. Similarly, $k^{-}$ are all the negative pixel embeddings that have different label than $f_{i}$. All the
embeddings in the loss function are $L_{2}$ normalized. For a single image sample, we get the final contrastive loss by averaging over all the embeddings:

\begin{align*}
    L_{CON} = \frac{1}{k^2}\sum_{i \in k^{2}}L_{i}
\end{align*}

Our final loss to optimize then becomes:

\begin{align*}
    L = L_{CE} + L_{CON}
\end{align*}

Here, $L_{CE}$ is the corss-entropy loss.

\section{Experiments}
In this section, we describe experiments on three different manipulation datasets to explore the effectiveness of CFL-Net. These datasets are general manipulation datasets containing several manipulation types and are not specific to only a single manipulation type. The evaluation metric we use is pixel-wise Area Under Curve (AUC) score \cite{hu2020span}.

\subsection{Datasets}
\begin{itemize}
    \item \textbf{IMD-20 \cite{novozamsky2020imd2020}} is a real-life manipulation dataset made by unknown people and collected from the Internet. Hence, this dataset contains various types of manipulations. There are a total of 2010 image samples in the dataset.
    
    \item \textbf{CASIA \cite{dong2013casia}} CASIAv2 contains 5123 images and CASIAv1 contains 921 images. Samples from this dataset are manipulated by splicing and copy-move forgery. Also, image enhancement techniques including filtering and blurring are applied to the samples for post-processing.
    
    \item \textbf{NIST-16 \cite{nist}} contains 584 image samples with ground-truth masks. Samples from NIST16 are manipulated by splicing, copy-move and removal, and are post-processed to hide visible traces.
    
\end{itemize}

For each dataset, we use the same procedure as \cite{hao2021transforensics} for train-val-test splits. It should be noted that, previous methods such as, \cite{hu2020span, bappy2019hybrid, kwon2021cat} usually pre-train their models on large ($\approx$1M samples) synthetic manipulation datasets and then fine-tune the models on the datasets mentioned above to report the final result. However, in this paper, to evaluate solely the model's performance, we do not create a synthetic manipulation dataset to pretrain our model. Interestingly, without taking a resort to any large synthetic manipulation dataset, our model outperforms the baseline models.


\subsection{Implementation Details}
We use ResNet-50 as encoder for both the streams. 
We train CFL-Net with Adam optimizer with a learning rate of 1e-4. We reduce the learning rate by $20\%$ after each 20 epochs. We resize the input images to $256 \times 256$. We divide $F$ into a total of $64 \times 64$ patches. The temperature $\tau$ of \eqref{con_loss_final} is set as 0.1. Cross-entropy loss is weighted to give the tampered class ten times more weight. We set the batch size to 4 and train the model on NVIDIA RTX Titan GPU over 100 epochs.

\subsection{Baseline Models}
We compare our method with various baseline models, which are described below:
\begin{itemize}
    \item J-LSTM \cite{bappy2017exploiting} employs a hybrid CNN-LSTM architecture to capture the discriminative features between untampered and tampered regions.
    \item RGB-N \cite{zhou2018learning} adopts a two stream parallel network to separately discover tampering features.
    \item ManTraNet \cite{wu2019mantra} uses a feature extractor to capture the manipulation traces and a local anomaly detection network to localize the manipulated regions.
    \item SPAN \cite{hu2020span} uses a pyramid architecture to and self-attention blocks to model the dependency of image patches.
    \item Transforensics \cite{hao2021transforensics} uses vision transformers with dense self-attention encoders and dense correction modules to model all pairwise interactions between local patches at different scales.
\end{itemize}

\section{Results}
In this section we report the results of our experiments. We divide the result section into two subsections in order to show the quantitative and qualitative results separately. We also perform ablation study. 
\subsection{Quantitative Analysis}
\begin{table}[]
\begin{tabular}{lccc}
\hline
Methods               & \multicolumn{1}{l}{NIST} & \multicolumn{1}{l}{CASIA} & \multicolumn{1}{l}{IMD-20} \\ \hline
J-LSTM (ICCV'17)         & -                        & -                         & 48.7                    \\
RGB-N (CVPR'18)          & 93.7                     & 79.5                      & -                       \\
Mantranet (CVPR'19)      & 79.5                     & 81.7                      & 81.3                    \\
SPAN (ECCV'20)           & 96.1                     & 83.8                      & -                    \\
Transforensics (ICCV'21) & -                     & 85.0                      & 84.8                    \\
Ours                  & \textbf{99.7}            & \textbf{86.3}                      & \textbf{89.9}           \\ \hline
\end{tabular}
\caption{AUC Scores (in \%).}
\label{table:auc}
\end{table}

We report the AUC scores (in \%) of our method and the baseline models in Table \ref{table:auc}. It should be noted that the results of RGB-N and SPAN stated here are the fine-tuned results as reported in their respective papers. J-LSTM and Transforensics do not perform any pre-training. Although ManTraNet pre-trains their model on synthetic manipulation dataset, they do not fine-tune on specific dataset. Looking at the table, it can be seen that CFL-Net achieves the best localization performance on all the datasets amongst the baseline models. Especially, CFL-Net outperforms all the baseline models by a big margin on IMD-20 dataset, which is a real-life manipulation dataset with various forgery types. Specifically, CFL-Net achieves an AUC score of $89.9\%$ on IMD-20 dataset, which is a $5.1\%$ improvement over the second most well-performing model - Transforensics. Hence, it validates our claim that CFL-Net is well-suited to localize real-life forgery. Our model also outperforms baseline models on the rest of the datasets - Casia and Nist. Moreover, it is worth pointing out that CFL-Net achieves these results without pre-training on synthetic manipulation data.

We argued that, in consequence of adding contrastive loss, our proposed model does not focus on specific forgery footprints but learns more generalized features. Hence, our model should generalize better across different manipulation datasets than the model trained without contrastive loss. For this reason, in our next experiment, to get an idea of how well our proposed method generalizes across datasets, we evaluate the models trained on one dataset and evaluate on the test sets of the remaining datasets. 

\begin{table}[]
\centering
\begin{tabular}{c|c|ccc}
\hline
Datasets                &     & NIST          & CASIA         & IMD-20        \\ \hline
\multirow{2}{*}{NIST}   & w/o & 98.3          & 67.1          & 66.4          \\ \cline{2-5} 
                        & w   & \textbf{99.7} & \textbf{67.6} & \textbf{69.8} \\ \cline{1-5}
\multirow{2}{*}{CASIA}  & w/o & 79.3          & 84.9          & 75.5          \\ \cline{2-5}
                        & w   & \textbf{79.9} & \textbf{86.3} & \textbf{77.8} \\ \cline{1-5}
\multirow{2}{*}{IMD-20} & w/o & 74.37         & 74.1          & 85.2          \\ \cline{2-5}
                        & w   & \textbf{91.8} & \textbf{75.6} & \textbf{89.9} \\ \hline
\end{tabular}
\caption{The left-most column shows the datasets models are trained on. The later columns are the datasets where the models are evaluated on. 'w/o' - CFL-Net trained without contrastive loss, 'w' - CFL-Net trained with contrastive loss. Results are in \% AUC.}
\label{table:gen}
\end{table}

\begin{figure*}[t!]
\setlength\tabcolsep{1pt}
\centering
\begin{tabular}{ccccc}
 \includegraphics[width=1.5in]{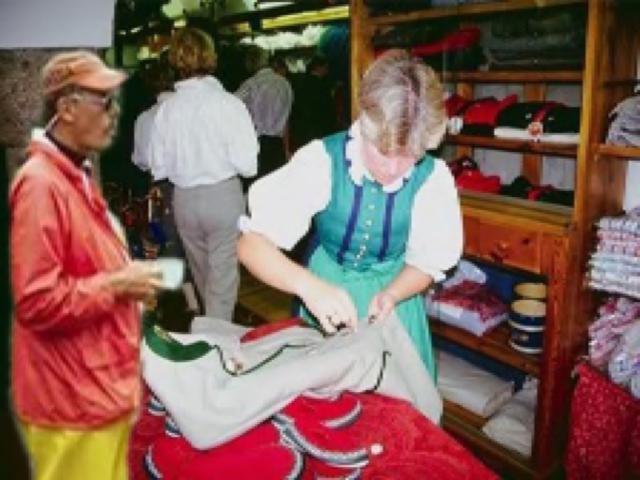} &
 \includegraphics[width=1.5in]{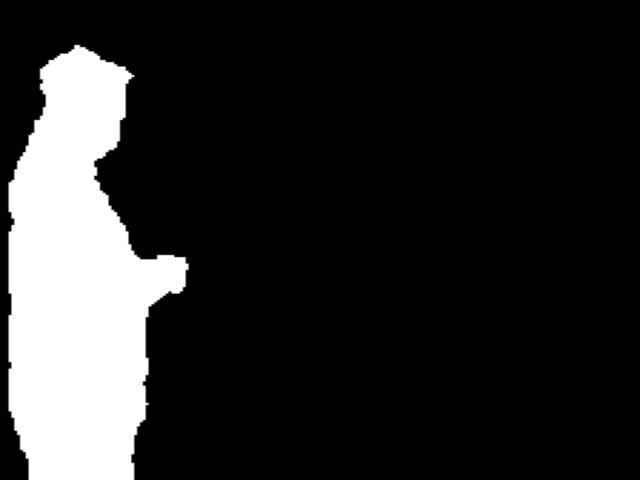} &
 \includegraphics[width=1.5in]{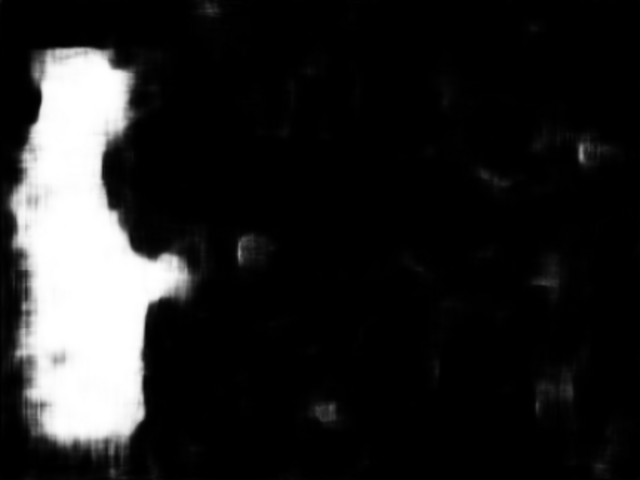} &
 \includegraphics[width=1.5in]{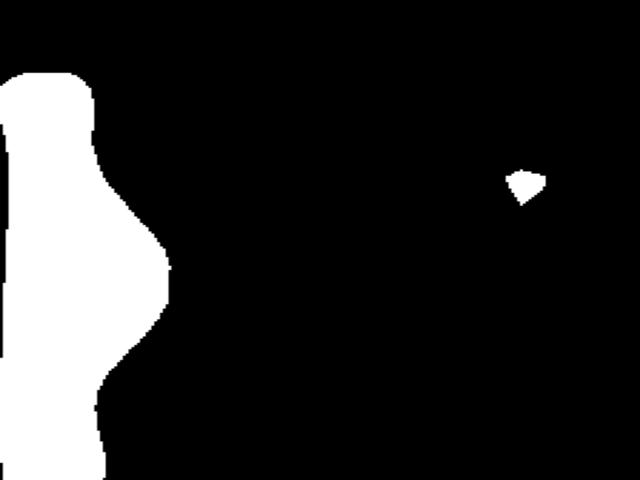} \\
 \includegraphics[width=1.5in]{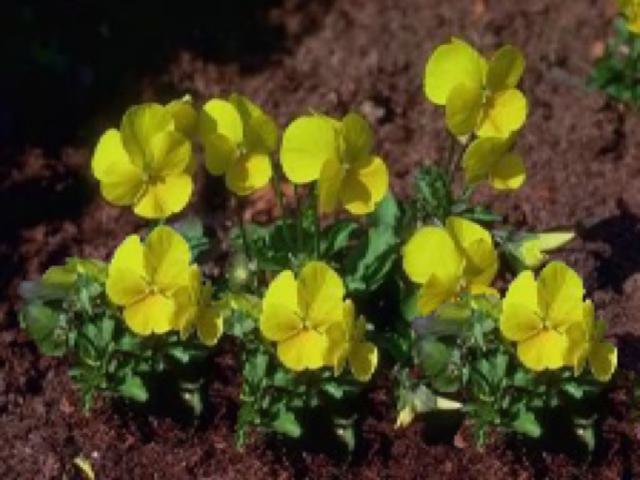} &
 \includegraphics[width=1.5in]{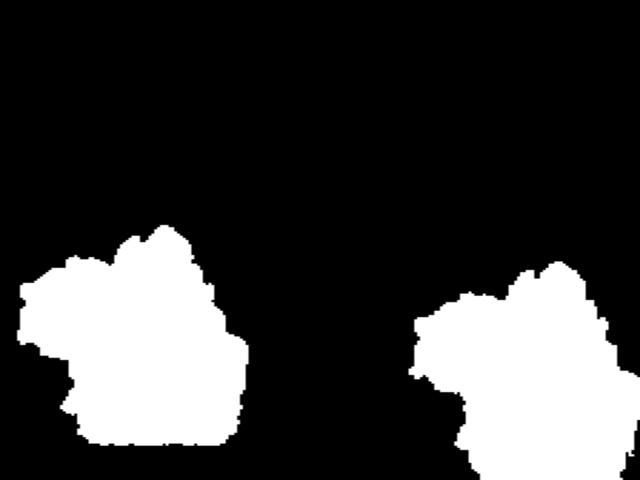} &
 \includegraphics[width=1.5in]{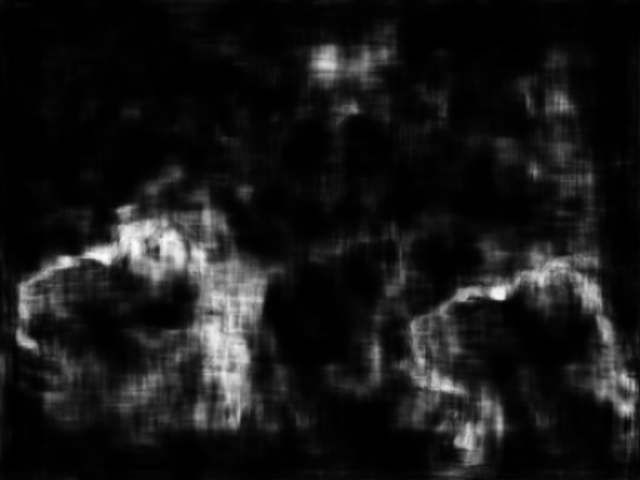} &
 \includegraphics[width=1.5in]{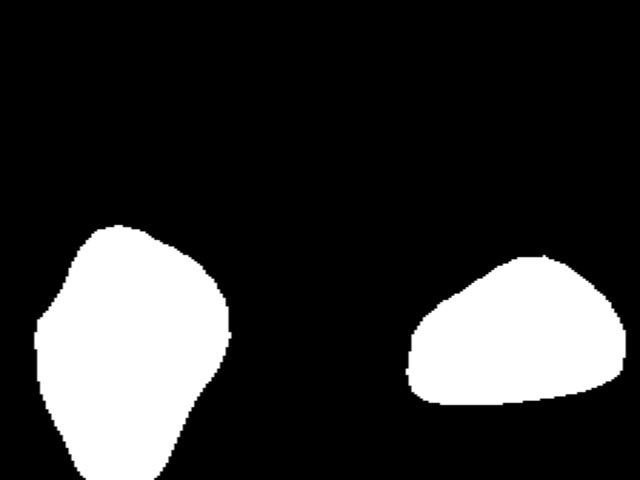} \\
 \includegraphics[width=1.5in]{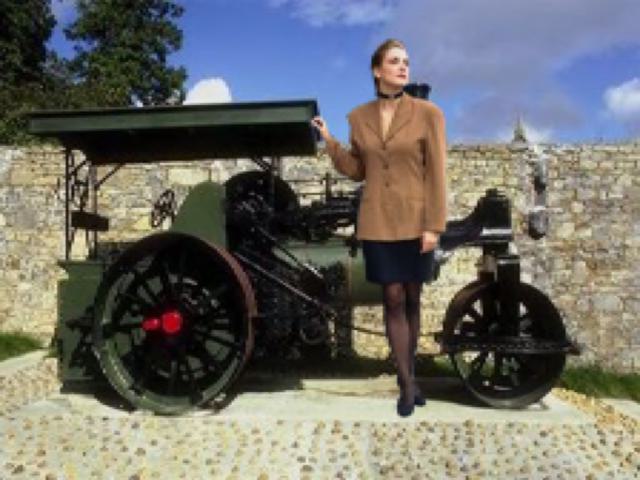} &
 \includegraphics[width=1.5in]{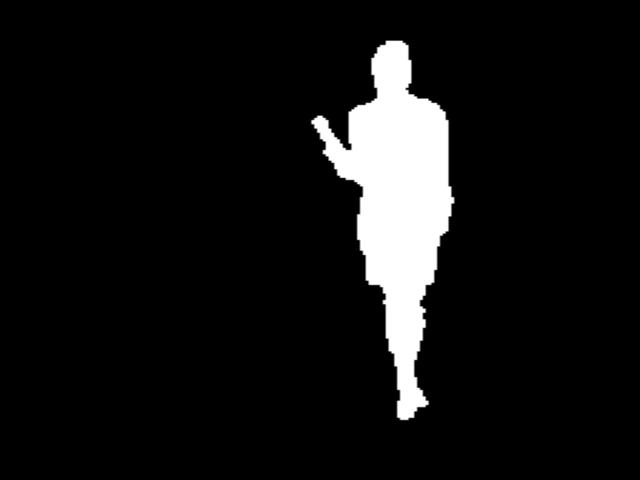} &
 \includegraphics[width=1.5in]{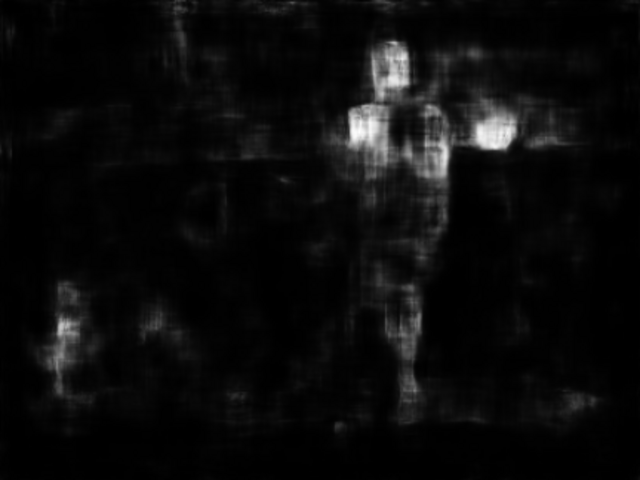} &
 \includegraphics[width=1.5in]{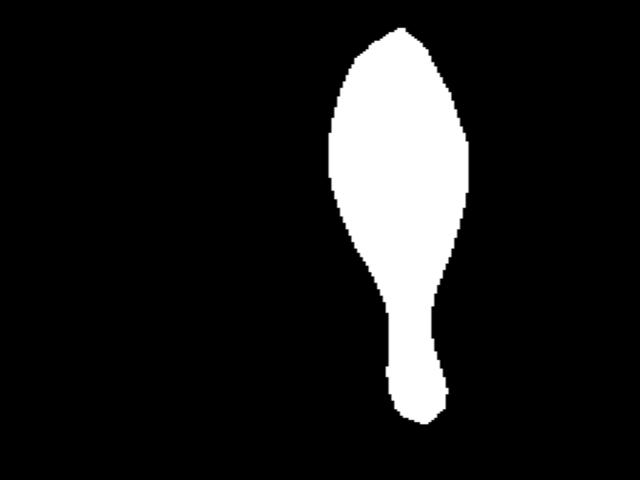}  \\
 \includegraphics[width=1.5in]{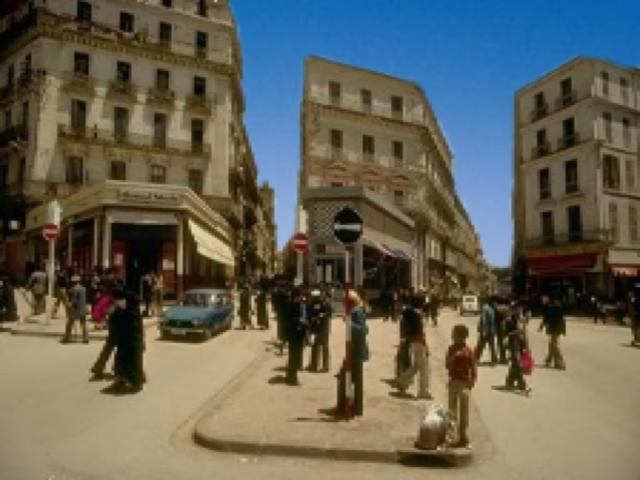} &
 \includegraphics[width=1.5in]{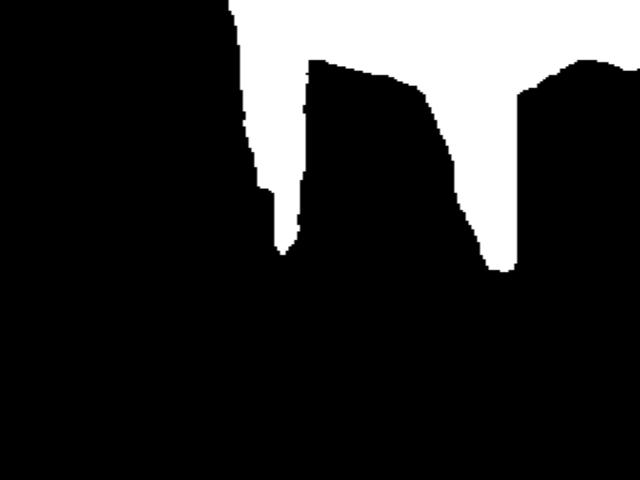} &
 \includegraphics[width=1.5in]{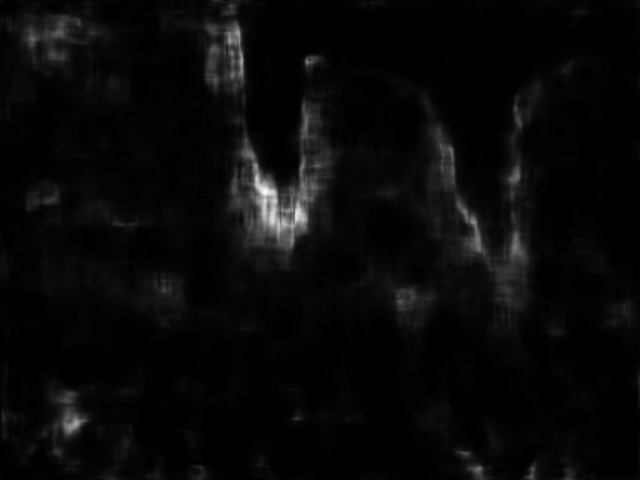} &
 \includegraphics[width=1.5in]{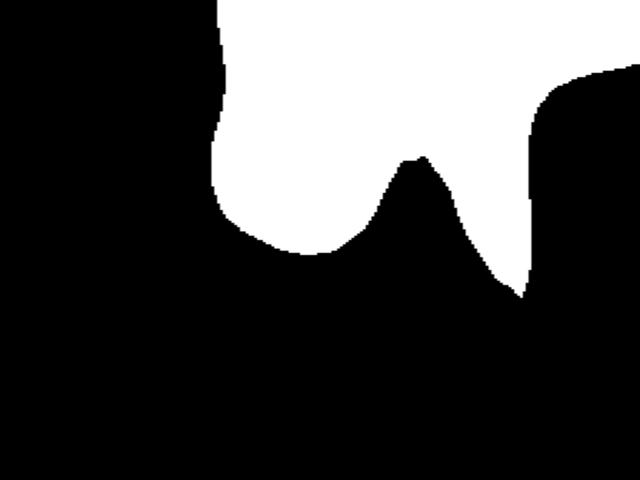} \\
 \includegraphics[width=1.5in]{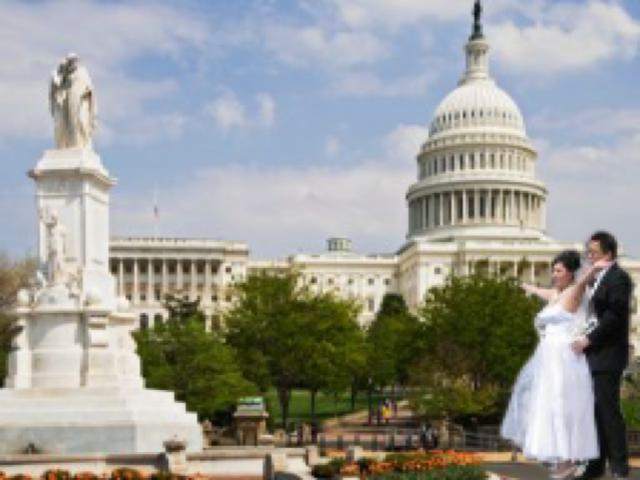} &
 \includegraphics[width=1.5in]{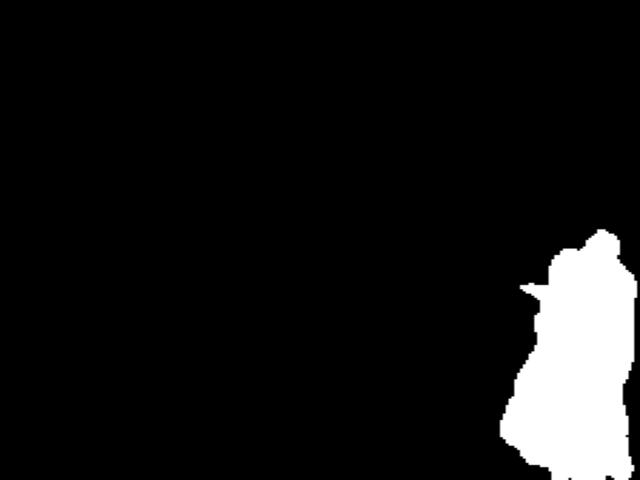} &
 \includegraphics[width=1.5in]{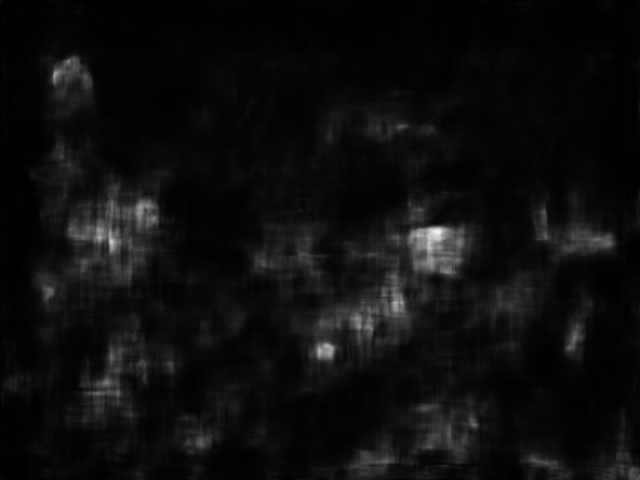} &
 \includegraphics[width=1.5in]{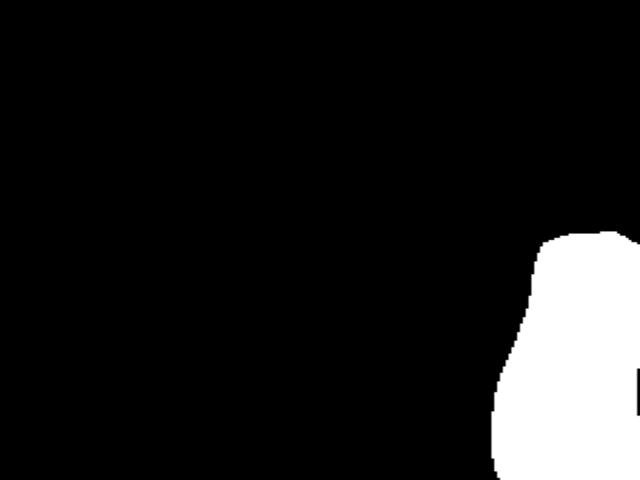} \\
 \textbf{\footnotesize{Manipulated Image}} &
 \textbf{\footnotesize{GT Mask}} &
 \textbf{\footnotesize{ManTraNet Prediction}} &
 \textbf{\footnotesize{CFL-Net Prediction}} \\
\end{tabular}
\caption{Comparison of the predicted mask with ManTraNet. It is evident that prediction of CFL-Net is closer to the GT mask compared to ManTraNet.}
\label{fig:qual}
\end{figure*}


Table \ref{table:gen} shows the results. It is evident that CFL-Net trained with contrastive loss performs very well in generalizing across datasets. In all the cases this model performs better than the model trained without the contrastive loss. When trained on IMD-20 and evaluated on the test set of NIST, our proposed model even outperforms the AUC score of ManTraNet. The most performance boosts are seen when trained on IMD-20 dataset. IMD-20 is the real-life image manipulation dataset and hence training on this dataset helps the model learn most generalizable features. Hence our proposed model trained on IMD-20 and evaluated on rest of the datasets yields the most performance improvement over the model trained without contrastive loss.

It should also be noted that both models trained on NIST and evaluated on the other datasets perform poorly because NIST has very few images, i.e., 584 images in the dataset. Hence, it is difficult to generalize to other datasets using NIST. Still, our proposed model managed to perform better than the model trained without contrastive loss.  

\begin{figure}[h!]
\begin{tabular}{cc}
\subfloat[CE Loss (IMD-20)]{\includegraphics[width = 1.5in]{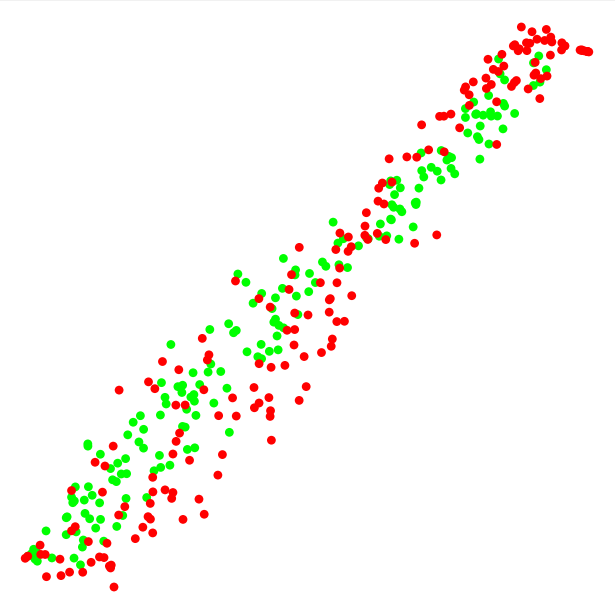}} &
\subfloat[CE + CON Loss (IMD-20)]{\includegraphics[width = 1.5in]{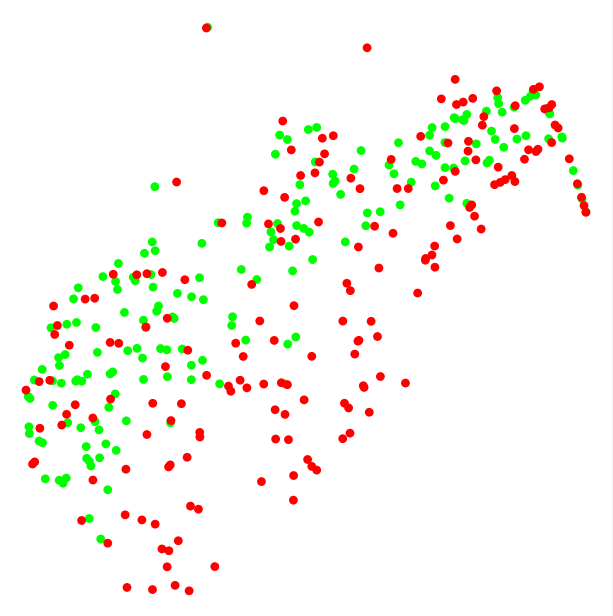}} \\
\subfloat[CE Loss (CASIA)]{\includegraphics[width = 1.5in]{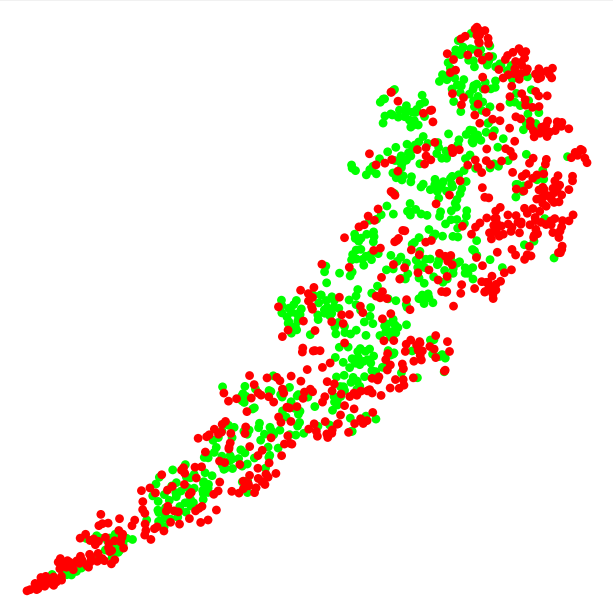}} &
\subfloat[CE + CON Loss (CASIA)]{\includegraphics[width = 1.5in]{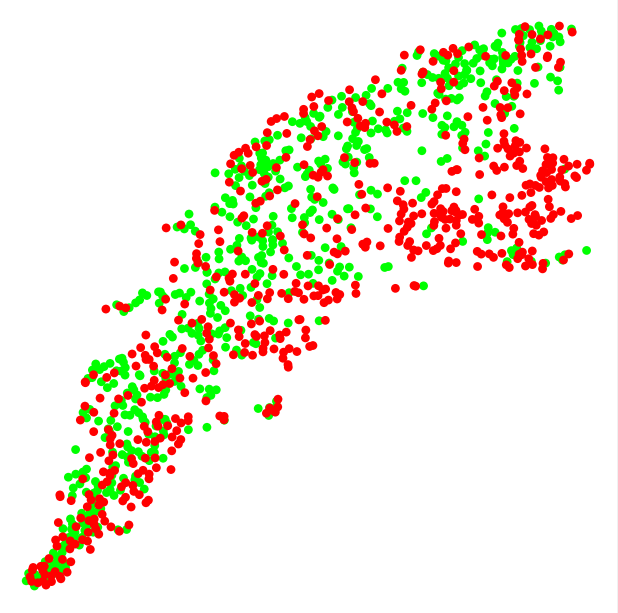}}
\end{tabular}
\caption{Left column shows t-SNE diagram of the mean features on IMD-20 and CASIA testsets when CFL-Net is trained using only cross-entropy loss. Right column corresponds to CFL-Net trained using both cross-entropy loss and contrastive loss. Green = Untampered feature, Red = Tampered feature.}
\label{fig:vis}
\end{figure}

\subsection{Qualitative Analysis}
Here we visualize a few of the predicted masks from the test sets. We also show the corresponding predicted mask of ManTraNet \cite{wu2019mantra} for comparison against our CFL-Net. ManTraNet's implementation and the saved model are made publicly available by the authors, which we employ here for the experiment. The results are shown in Figure \ref{fig:qual}. From the figure, it is evident that masks predicted by CFL-Net are closer to the ground truth masks. On the other hand, ManTraNet struggled to detect the manipulated region in most of the cases.

Next, in order to show that our contrastive loss preserves the feature variations by avoiding clustering of same class features, we visualize via t-SNE the class features obtained from the segmentation head in Figure \ref{fig:vis}. The left column shows the mean feature vectors per image sample on IMD-20 and CASIA test sets when CFL-Net is trained using only cross-entropy loss. Visibly, the features corresponding to both untampered (green color in figure) and tampered (red color in figure) regions are congested here. On the other hand, the right column shows the mean features when CFL-Net is trained using both cross-entropy and contrastive loss. Here, the features corresponding to both regions are more dispersed. Hence, different manipulation footprints are more separable. This experiment demonstrates that the traditional cross-entropy loss reduces generalization in case of image forgery localization due to the intra-category invariance, while our proposed method can improve the generalization by diverging the feature distribution.

\subsection{Ablation Study}
In this subsection, we conduct ablation experiments to
study how the proposed loss of CFL-Net influence the localization performance. Specifically, we train CFL-Net without the contrastive loss and then report the results to get an idea of the influence of contrastive loss.

\begin{table}[h!]
\centering
\begin{tabular}{lccc}
\hline
Methods                  & \multicolumn{1}{l}{NIST} & \multicolumn{1}{l}{CASIA} & \multicolumn{1}{l}{IMD} \\ \hline
CE Loss & 98.3                     & 84.9                      & 85.2                    \\
CE + CON Loss    & \textbf{99.7}            & \textbf{86.3}             & \textbf{89.9}        
\\ \hline
\end{tabular}
\caption{AUC scores (in \%) for CFL-Net trained with different loss settings. CE = Cross-entropy loss, CON = Contrastive Loss.}
\label{table:ablation}
\end{table}

In Table \ref{table:ablation} we report the results. It is clear from the table that adding the contrastive loss indeed helps in localization. The improvement is much more prominent on the real-life image manipulation dataset IMD-20. Contrastive loss helps to improve the AUC score by $4.7\%$. It should be noted that without the contrastive loss our method already achieves very good results. The reason is that our model is similar to RGB-N \cite{zhou2018learning} in the regard that we also use two stream network, i.e, RGB and SRM streams. In addition, we carefully supplement our network with ASPP module and Deeplab decoder head, which helps to improve the overall performance compared to RGB-N. Using contrastive loss further improves our results and helps to outperform all the other baseline models.

\section{Conclusion}
In this paper, we approached the general-purpose image forgery localization problem from a new perspective, i.e., using contrastive learning. We identified a major drawback of existing methods that focus on specific forgery footprints and use cross-entropy loss without any constraints to localize forgery. To address the drawbacks, we supplemented cross-entropy loss with contrastive loss and proposed a novel image forgery localization method named \textit{Contrastive Forgery Localization Network} or \textit{CFL-Net}. We conducted experiments on three benchmark image manipulation datasets and compared our results with major forgery localization methods of recent years. CFL-Net outperformed all the methods in terms of AUC metric. Moreover, the improvement is much more prominent on the real-life image manipulation dataset IMD-2020. Amongst the future works, a more sophisticated fusing mechanism can be considered to fuse the feature maps from the RGB and SRM streams. For example, attention modules or recently proposed vision transformers can be employed as a fusion mechanism.

\section{Acknowledgment}
This work was supported by ICT Division - Government of Bangladesh and Independent University Bangladesh (IUB). In addition, this work was partly supported by the Basic Science Research Program through National Research Foundation of Korea (NRF) grant funded by the Korean Ministry of Science and ICT (MSIT) under No. 2020R1C1C1006004 and Institute for Information \& communication Technology Planning \& evaluation (IITP) grants funded by the Korean MSIT: (No. 2022-0-01199,  Graduate School of Convergence Security at Sungkyunkwan University), (No. 2022-0-01045, Self-directed Multi-Modal Intelligence for solving unknown, open domain problems), (No. 2022-0-00688, AI Platform to Fully Adapt and Reflect Privacy-Policy Changes), (No. 2021-0-02068, Artificial Intelligence Innovation Hub), (No. 2019-0-00421, AI Graduate School Support Program at Sungkyunkwan University), and (No. 2021-0-02309, Object Detection Research under Low Quality Video Condition).

{\small
\bibliographystyle{ieee_fullname}
\bibliography{egbib}
}

\end{document}